\documentclass{article}

\usepackage[final,nonatbib]{nips_2017}

\usepackage[utf8]{inputenc} 
\usepackage[T1]{fontenc}    
\usepackage{hyperref}       
\usepackage{url}            
\usepackage{booktabs}       
\usepackage{amsfonts}       
\usepackage{nicefrac}       
\usepackage{microtype}      
\usepackage{amsmath}
\usepackage{algorithm, algorithmic}
\usepackage{graphicx}
\usepackage{enumitem}
\graphicspath{ {images/} }

\title{A Spatial Mapping Algorithm with Applications in Deep Learning-Based Structure Classification}

\author{
  Thomas Corcoran \\
  Lawrence Berkeley National Laboratory \\
  Hood College of Frederick Maryland \\
  \texttt{tcorcoran@lbl.gov} \\
  \And
  Rafael Zamora-Resendiz \\
  Lawrence Berkeley National Laboratory \\
  Hood College of Frederick Maryland \\
  \texttt{rzamoraresendiz@lbl.gov} \\
  \And
  Xinlian Liu\\
  Lawrence Berkeley National Laboratory\\
  Hood College of Frederick Maryland \\
  \texttt{xinlianliu@lbl.gov} \\
  \And 
  Silvia Crivelli \\
  Lawrence Berkeley National Laboratory \\
  \texttt{sncriveli@lbl.gov} \\
}

\begin{document}

\maketitle

\begin{abstract}
 Convolutional Neural Network (CNN)-based machine learning systems have made breakthroughs in feature extraction and image recognition tasks in two dimensions (2D). Although there is significant ongoing work to apply CNN technology to domains involving complex 3D data, the success of such efforts has been constrained, in part, by limitations in data representation techniques. Most current approaches rely upon low-resolution 3D models, strategic limitation of scope in the 3D space, or the application of lossy projection techniques to allow for the use of 2D CNNs. To address this issue, we present a mapping algorithm that converts 3D structures to 2D and 1D data grids by mapping a traversal of a 3D space-filling curve to the traversal of corresponding 2D and 1D curves. We explore the performance of 2D and 1D CNNs trained on data encoded with our method versus comparable volumetric CNNs operating upon raw 3D data from a popular benchmarking dataset. Our experiments demonstrate that both 2D and 1D representations of 3D data generated via our method preserve a significant proportion of the 3D data's features in forms learnable by CNNs. Furthermore, we demonstrate that our method of encoding 3D data into lower-dimensional representations allows for decreased CNN training time cost, increased original 3D model rendering resolutions, and supports increased numbers of data channels when compared to purely volumetric approaches. This demonstration is accomplished in the context of a structural biology classification task wherein we train 3D, 2D, and 1D CNNs on examples of two homologous branches within the Ras protein family. The essential contribution of this paper is the introduction of a dimensionality-reduction method that may ease the application of powerful deep learning tools to domains characterized by complex structural data.
 
\end{abstract}

\section{Introduction and Motivation}

Deep learning has emerged as an effective tool in applications that contain classification and feature extraction components \cite{lecun15}. The design of convolutional neural networks (CNNs) was inspired in part by biological vision processing mechanisms, as well as by existing artificial image processing systems. In particular, such networks were originally intended to operate upon data containing a known grid-like topology, a property which most commonly manifests as ordinary RGB or greyscale images \cite{Goodfellow-et-al-2016}. CNNs learn abstract hierarchies of features found within their inputs by convolving large numbers of randomly-initialized feature detectors, called 'kernels', over their input space. These feature-detecting kernels are sensitive to various combinations of pixel clusters, and activate when they detect the presence of their encoded feature in a given patch of the input. Each subsequent hidden layer in a CNN performs convolutions on increasingly abstracted representations of the input data. Kernels in early, lower, layers detect low-level features such as curves and edges, while kernels found in later, higher, layers detect high-level features such as complex shapes and textures. This relationship is intuitive, since high-level features are naturally composed of many lower-level features which, in turn, are detected earlier in the network. In the best-case scenario, the final, highly-distilled features recognized by such a network represent the archetypal characteristics of a given class, enabling an efficient and general classification capability for data points of that class.

Figure 1 contains a limited overview of the current landscape of problem domains that exist in varying dimensions and the various deep learning approaches that have been developed to address them. While 2D and 1D CNN technology has at this point become well-tuned to addressing typical 1D and 2D problem domains, the process of mastering the application of deep learning tools against 3D and higher-dimensional data remains ongoing. 

In recent years 2D CNN technology has advanced rapidly due to its successes in practical settings, such as in natural image recognition tasks, with new network architectures and training mechanisms routinely redefining state-of-the-art computer vision performance standards since AlexNet's groundbreaking entry into the 2012 ImageNet challenge \cite{krizhevsky2012imagenet}. The prominent role played by 2D CNNs in popular applications that make use of facial recognition (e.g., Facebook's automatic photo tagging) may in part explain the explosive growth of this technology's sophistication.

Although perhaps less discussed than their 2D counterparts, 1D CNNs have also advanced significantly in recent years thanks to the same factors that have fueled the current deep learning and AI "summer", namely the availability of huge amounts of data and compute resources. In fact, the successes of 2D CNNs in the realm of image recognition tasks have almost been matched by 1D CNNs in their own niche of natural language processing. Recent work has explored the frontiers of 1D CNN applications by applying the technology against amino acid sequence data \cite{DBLP:journals/corr/HouAC17}, but although this work is promising, it appears to still view the applicability of 1D CNN technology through the lens of its traditional role as a sequence-based learning tool. 

\begin{figure}[h]
    \centering
    \includegraphics[width=10cm]{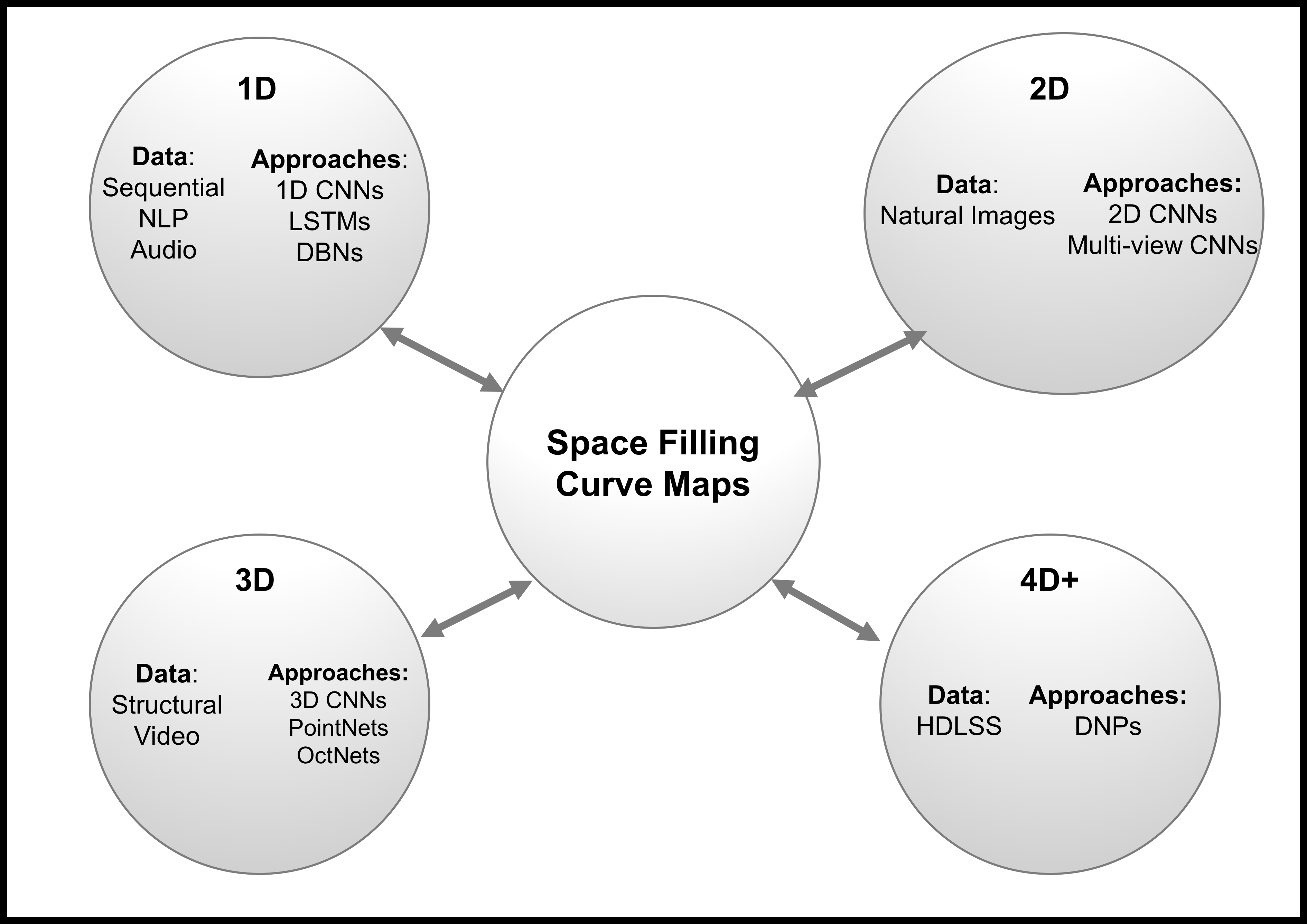}
    \caption{Various deep learning approaches have been developed in order to address datasets in different dimensions. Our space-filling curve-based mapping algorithm serves as the bridge between different deep learning approaches by providing a dimensionality conversion method that may help to expand the boundaries of deep learning applications into the realm of sparse, high dimension data such as that found in bioinformatics.}
\end{figure}

Meanwhile, 3D CNNs have shown promise in the classification of spatio-temporal data \cite{karpathy2014large}, as well as in the classification of discrete 3D structural data \cite{maturana2015voxnet}. However, despite recent work done to optimize 3D CNNs for improved performance on large, sparse datasets (which represent the majority of structural datasets) in projects such as OctNet \cite{riegler2016octnet} or Sparse 3D CNNs \cite{DBLP:journals/corr/NotchenkoKB16}, the performance of 3D CNNs is still limited in several critical ways when compared against their 2D counterparts. In particular, the 3D convolution operation suffers from a cubic time complexity (an order higher than in the 2D case), a property that has led many researchers to choose a $30^3$ rendering size for their 3D CNN work \cite{FPNN16}. While this resolution may be sufficient for certain types of coarse recognition tasks, such a constraint poses a real challenge to the prospects of applying 3D CNN technology to domains requiring a more intricate, higher-resolution treatment of data, such as in the case of structural biology, where having the option to use a highly-detailed rendering of relevant biomolecules without needing to drastically limit the region of consideration is desirable.

A related issue that hampers 3D CNN use in complex structural or spatio-temporal analysis tasks is that the computational costs associated with training such networks scale poorly as the networks are provided with additional data channels (i.e., as the richness of the input data is increased). This is a severe limitation that seems to relegate pure 3D CNN approaches to the realm of relatively simple object recognition tasks where geometric information alone is sufficient to learn from - an observation that is supported by the relatively low-resolution and mono-channeled datasets that currently dominate 3D object benchmarking communities, such as ShapeNet \cite{shapenet2015} and ModelNet \cite{ModelNet15}. In order to work around these limitations, current approaches have focused on the creative use of low-resolution volumetric systems that ingest subsets of large 3D models, and also on the use of so-called "multi-view CNNs" that operate by learning from many 2D pictures of 3D objects taken from various angles \cite{2016arXiv160403265Q} \cite{AtomNet}. However, neither of these approaches fully address the aforementioned limitations and, critically, neither offers the potential for easing the application of existing machine learning technology against datasets containing more than three dimensions. In other words, such approaches are incomplete and are unlikely to scale. 

Recently, progress has been announced by Liu et al. on applying deep neural networks to high dimension, low sample size (HDLSS) data such as that found in bioinfomatics through the use of their Deep Neural Pursuit system \cite{ijcai2017-318}. This system has shown promising results against very high-dimensional datasets, but it remains to be seen how effective it is against data composed of locally-correlated features such as those addressed by CNNs (e.g., "structural" problems).  

At least in the case of 1D, 2D, and 3D data types, the application of deep learning technology to various problem domains has followed a predictable pattern as far as the pairing of CNN dimensionality and domain dimensionality are concerned. It seems that, for the most part, problems for which data exist in 2D have been addressed by 2D deep learning systems, and the same for 3D and 1D cases as well. This trend seems especially clear in the case of structural (i.e., 3D) data, for which there currently exist only a small number of practical deep learning approaches, none of which appear to be optimal, possibly due to the aforementioned performance limitations of current 3D CNN technology (or simply current compute capacity). 

Although this trend is in some ways quite intuitive, another explanation underlying it may be the general difficulty associated with the process of dimensionality reduction in the context of deep learning. In other words, since CNNs operate upon the assumption that the important features of their input data manifest to some degree as clustered semantically-linked regions, it is vital that any dimensionality reduction scheme preserve these clusters to the greatest extent possible in order to allow the lower-dimension CNN to learn from them. The development of such a mapping scheme is complicated by all of the same issues that bedevil any dimensionality-reduction process, most notably information loss. 

In this paper, we demonstrate a novel space-filling curve-based transformation algorithm which can map high-dimension data grids to lower dimensions while preserving enough locality information to be exploited by CNNs operating in the lower dimension. Specifically, we demonstrate the generation of large 2D and 1D datasets from 3D structural data and the subsequent training of 2D and 1D CNNs upon said data. We show that our mapping of 3D structures to lower-dimension representations can be used to train CNNs that perform comparably to 3D CNNs on 3D shape recognition tasks, with the specific task example used being Princeton’s ModelNet10 shape recognition benchmark \cite{wu20153d}. After demonstrating the potency of our approach on the ModelNet10 benchmarking dataset, we proceed to an exploration of its characteristics through the lens of a structural biology classification task, namely the discrimination between two homologous branches of the Ras protein family. Through this example, we show that our methodology reaches a high classification performance standard while offering the potential for significantly reduced computational costs compared to 3D CNNs, all while supporting the addition of large numbers of additional data channels beyond what is currently feasible for purely volumetric approaches. This last point in particular means that our methodology potentially opens a new pathway for the application of deep learning techniques to problem domains containing complex structural problems that require high-resolution, multi-channeled data representations to enable learning of the potentially quite subtle features that may characterize classes in the domain. 

\section{Methodology}

\subsection{Underlying Principles}

A space-filling curve (more precisely, a \textit{plane-filling function}) is a linear mapping of a multi-dimensional space, and the typical goal underlying the application of such a curve to real-world problem domains is to impose an absolute ordering upon the multi-dimensional space in such a manner as to preserve a maximum of spatial locality between elements of that space \cite{moon01}. Such curves have been successfully applied to tasks in data organization, data compression, image half-toning and color quantization \cite{lindenbaum96}. However, it is important to note that these applications all involve the mapping of $M$-dimensional data to 1 dimension, and that the method we propose here is more general, allowing for the mapping of $M$-dimensional data to an arbitrary target dimension. 

Since CNN operations are based upon an intelligent exploitation of locally-correlated semantic clusters present within their input data, the use of space-filling curve mappings as a preprocessing tool to enable CNN training on sparse, high-dimensional data is natural. The concept of a space-filling curve-based dimensionality reduction scheme arose as the synthesis of several observations regarding the operation of traditional CNNs and a list of desired characteristics for a new data representation. Among the more important characteristics that we sought were that the transformation be reversible (i.e., bijective) in order to facilitate visualization of interpretable saliency maps, that it be based upon well-understood principles, and that it enable the transformation of high-dimensional data into various lower-dimension spaces if possible. 

An item of particular contextual relevance that illuminated a space-filling curve-based approach as potentially viable is the fact that traditional CNNs rely upon a relatively crude and ineffective mechanism in order to learn high-level spatial relationships between the features present in their input data. This mechanism, called max pooling, is not robust against simple transformations of features and is not capable of reliably learning the hierarchical structures that compose complex data. Instead, CNNs employing max pooling rely upon a strong ability to detect relatively small features without regard to their global arrangement. This is a significant limitation that the computer vision community has been aware of for some time and it was a primary motivating factor behind the development of Geoffrey Hinton's new Capsule Networks \cite{2017arXiv171009829S}. However, capsule technology is new and, as of this writing in early 2018, still in need of efficiency improvements before it can be practically applied to large datasets. Any dimensionality reduction method risks losing important information, and one issue with a space-filling curve-based mapping approach in particular is that it will invariably fail to preserve the global arrangement of features present in a given data instance when mapping that data into a space of different dimensionality. Given the pseudo-invariance of traditional CNNs to the global arrangement of features, however, this characteristic of a space-filling curve-based mapping approach seemed surmountable. 

\subsection{Dimensionality Reduction via Space-Filling Curve-Based Transformations}

\begin{figure}[h]
    \centering
    \includegraphics[width=6cm]{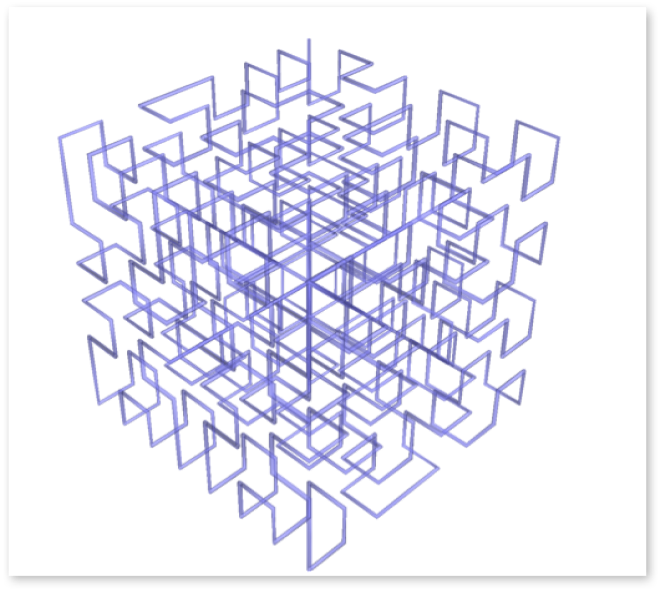}
    \includegraphics[width=6cm]{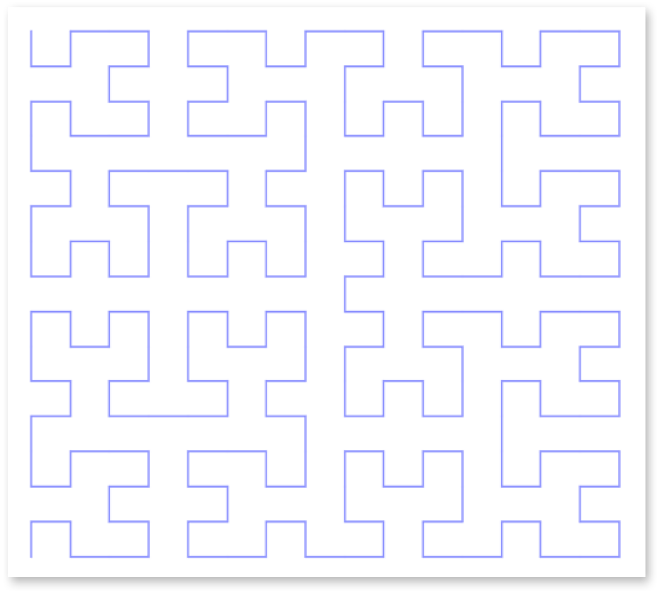}
    \caption{At left, a rendering of a Hilbert curve traversal of 3D space. At right, a Hilbert curve traversal of 2D space.}
\end{figure}

In the course of our work, we evaluated several different types of space-filling curves as potentially useful for mapping three-dimensional structural data into lower dimensions, with our ultimate goal being to utilize such a mapping as a preprocessing mechanism for the training of efficient one and two-dimensional convolutional neural networks. Among the different families of curves considered were the \textit{Z-order curve}, the \textit{Gray-coded curve}, and the \textit{Hilbert curve}. 

Since convolutional neural networks are only able to learn from data which are composed of clustered semantic regions (i.e. neighborhoods of semantically-linked points), and since it is assumed that any type of structural data of interest to researchers will be composed to some degree of such semantic clusters, it is important that a lower-dimensional representation of such data preserve as much neighborhood information as possible so that those features may be detectable by a network's convolutional filters. It is known that the Hilbert curve achieves a superior preservation of locality compared to alternative curves, and specifically when compared against both the Z-order and Gray-coded curves \cite{moon01} \cite{lindenbaum96}. Figure 2 visualizes a rendering of a Hilbert curve traversal of 3D and 2D spaces in order to hopefully provide a stronger intuition for our mapping process.

For the purpose of explaining our approach, we adopt the notational conventions and function descriptions found in \cite{lindenbaum96} as follows. 

\begin{itemize}
    \item Denote $[N]=\{1,...,N\}$. 
    \item A discrete $m$-dimensional space-filling curve of length $N^m$ is a bijective mapping denoted
    \begin{center}
    $C:[N^m] \rightarrow [N]^m$
    \end{center}
    such that 
    \begin{center}
    $d(C(i),\ C(i+1))=1\ \forall\ i \in [N^m - 1]$,    
    \end{center}
    where $d()$ is a Euclidian distance metric. 
\end{itemize}

The essential distillation of this mapping process is that a curve $C$ of length $N^m$ will traverse each of the $N^m$ points of an $m$-dimensional grid with side length $N$. It is true that, when operating in continuous space, plane-filling functions such as the Hilbert curve are not one-to-one but are in fact surjective functions \cite{tarver14}, and it has been shown that the Hilbert curve specifically is a continuous, surjective, and nowhere differentiable mapping \cite{sagan93}. However, since we are operating in the discrete, digital environment of a computer, we may treat the Hilbert curve transformation used in our encoding process as having a one-to-one correspondence between the relevant three-dimensional, two-dimensional, and one-dimensional spaces. 

This is directly relevant to our work since, for each voxel in the three-dimensional model space, there is exactly one value generated in the output one and two-dimensional representations of that space. In the case of two-dimensional representations, that value is a single pixel in a two-dimensional data grid (i.e., an image), and in the one-dimensional case that value manifests as a single entry in a $1 \times N$ vector, where $N^3$ is the total number of voxels used to describe the 3-dimensional space. It is the bijective nature of the discrete space Hilbert curve mapping that allows us to ensure the information-preserving characteristics of our dataset generation processes, as well as to generate attention map visualizations of regions of interest to the network, a process which will be described in forthcoming versions of this paper (see future work).  

By 'locality' we refer to the idea that proximity between points in $[N^m]$ is reflected in the curve traversal, a property which is almost optimally expressed in the Hilbert transposition \cite{lindenbaum96}. In other words, this property means that near neighbor indexes in the three-dimensional model space are also near neighbors within the one-dimensional space that results from the Hilbert curve traversal of the three-dimensional space. Perez \textit{et al.} quantify an average measure of locality preservation between points in $m$-dimensional space and their resulting positions in the 1-dimensional space as follows:

\begin{equation}
    L(C)\ = \sum_{i,j \in N^m,\ i<j}^{} \frac{|i-j|}{d(C(i),\ C(j))}.
\end{equation}

In order to map an $m$-dimensional space onto a space of dimensionality greater than one using the Hilbert curve traversal, we may take a composition of two separate bijective Hilbert-curve mappings, since the composition of bijections is itself a bijection \cite{BijectionLemma}. Then for two bijective mapping functions  $C_m:[N^m] \rightarrow [N]^m$ and $C_l:[N]^l \rightarrow [N^l]$, we have their bijective composition

\begin{align*}
    C_{m \rightarrow l}&=(C_m \cdot C_l)\\
    &=[N^m] \rightarrow [N^l].
\end{align*}

This function maps a linear traversal of a curve through an m-dimensional space to a linear traversal of a second curve of equivalent length through an l-dimensional space. Via a slight modification of Perez's $m$-dimension to one-dimension calculation, a similar measure of average locality preservation between the m-dimensional space and the l-dimensional space can be defined as follows:

\begin{equation}
    L(C_{(m \rightarrow l)})\ = \sum_{i,j \in [N^m],\ i<j}^{} \frac{|i-j|}{d(C_m(C_l(i)),\ C_m(C_l(j)))}.
\end{equation}

However, given that CNNs operate based upon neighborhoods of adjacent pixels within limited spatial ranges, a better metric to determine the locality preservation quality of the mappings in question can perhaps be provided by the following:

\begin{equation}
    L(C_{(m \rightarrow l)}, K_m, K_l)=\sum_{i,j\in N^m, i < j}^{}
    \begin{cases}
        1 & d(C_l(i),C_l(j)) \leq K_l \land d(C_m(i),C_m(j)) \leq K_m \\
        0
    \end{cases}
\end{equation}

where $m$ and $l$ stand for the side length of a given kernel within the respective spaces. This locality measure determines how well, on average, a mapping procedure preserves the set of points captured by a kernel of size $K_m$ in the original space within a region that may be encompassed by a kernel of size $K_l$ in the transformed space. Note that this measure of locality preservation makes no distinction between arrangements of elements within kernel regions - it is only sensitive to the absolute size of the preserved element set between spaces. 

\subsection{3D Model Rendering and Mapping to Specific Lower Dimensions}

Throughout this paper, all 3D objects are discretized into a binary 3D representation within a window space of shape $64^3$ using Patrick Min's binvox voxelizer software \cite{patrickmin}. Binvox and the related viewvox are relatively mature pieces of software and allow for easy manipulation and viewing of binvox data files, a requirement for working with the ModelNet10 benchmarking dataset. Although our methodology theoretically supports the generation and use of very large 3D model renderings (we have successfully tested model resolutions up to $256^3$, for example) we elected to fix our model resolutions at $64^3$ to ease the process of exploration and network training iteration. This resolution is convenient for multiple reasons. First, since it is nearly 8 times larger by voxel count than that used by many current 3D deep learning systems, it clearly demonstrates the improvements in model rendering resolutions that are possible by encoding data through our approach (a 3D CNN training on multi-channel $64^3$ models would eventually become quite costly, let alone at even higher resolutions). Second, this resolution means that each voxel in the 3D space corresponds to one cubic angstrom in real space. Each of the 262,144 voxels comprising the 3D models are mapped to individual pixels in the output 1D and 2D encodings, yielding 1 x 262,144 vectors and 512 x 512 pixel images, respectively (with each element corresponding directly to one voxel of the 3D space). 

Voxel occupancy is represented within the rendered 3D matrix using only a binary value for each channel of data describing the model. For instance, in the case of the ModelNet10 dataset described below, the only channel of information available to describe the 3D objects is geometric, and so there is only one binary occupancy value associated with each voxel. We use this simple model representation strategy to render 3D data for training 3D CNNs; the binary occupancy function and 3D representation strategy allow for a straightforward traversal by a Hilbert curve in order to generate 2D and 1D representations of the 3D data that form the datasets for our various CNN training comparisons. In order to format the structural data for 2D and 1D CNN training we generate each representation using the approaches detailed in algorithms 1 and 2.

\begin{algorithm}[h!]
  \caption{3D to 2D Hilbert Curve Mapping}
  \begin{algorithmic}
    \STATE Input: Matrix x of shape (64, 64, 64)
    \STATE Output: Matrix y of shape (512, 512)

    \STATE Hx $\gets$ 3D Hilbert curve of order 6
    \STATE Hy $\gets$ 2D Hilbert curve of order 9

    \STATE Procedures:
    \STATE $y$ $\gets$ empty matrix of shape (512, 512)
      \FOR{$i=1$ to length of $Hx$}
        \STATE $y[Hy[ i ]] \gets x[Hx[ i ]]$
      \ENDFOR
    \STATE return y  
  \end{algorithmic}
\end{algorithm}

\begin{algorithm}[h!]
  \caption{3D to 1D Hilbert Curve Mapping}
  \begin{algorithmic}
    \STATE Input: Matrix x of shape (64, 64, 64)
    \STATE Output: Array y of size (1, 262144)

    \STATE Hx $\gets$ 3D Hilbert curve of order 6

    \STATE Procedures:
    \STATE $y$ $\gets$ empty array of size (1, 262144)
      \FOR{$i=1$ to length of $Hx$}
        \STATE $y[ i ] \gets x[Hx[ i ]]$
      \ENDFOR
    \STATE return y  
  \end{algorithmic}
\end{algorithm}

\subsection{CNN Architectures}
Due to its structural nature, our data called for a somewhat unconventional approach to network hyperparameter selection, especially in the 2D case, where we found the design heuristics associated with classic 2D image recognition tasks (such as the ImageNet and MNIST competitions) to be entirely inadequate. Specifically, we found that the sparsity of 3D data in both its native and space-filling curve-encoded formats precludes the use of conventional 2D CNN kernel sizes. The scattering side-effect of our mapping algorithm means that important features of a given 3D class manifest in heavily-distorted and somewhat dispersed forms when encoded into 2D. We found that the most direct way to address this issue is through the use of relatively large kernel sizes in our 2D networks. This same principle holds true in the case of our 1D representations and CNNs as well, where large vector-shaped kernels were required to capture dispersed class features. 

For the purposes of exploring our mapping algorithm, and in order to facilitate a clear demonstration and comparison of results, we standardized our work around three different CNN variants for each of the 3D, 2D, and 1D cases. During this stage of our work, we opted to implement our own lightweight CNN architectures based upon the general outline of other successful volumetric and 2D designs instead of making use of previously-published architectures in order to avoid having to address some of the unique characteristics of those models. This issue is addressed in more detail in section 3.1 during the discussion of the ModelNet10 dataset and the results we obtained from training our networks on it. 

Each network configuration includes the same number of convolutional and pooling layers in order to enable observation of the effects caused by varying kernel sizes when learning from representations of different dimensionalities, since we believe that the question of what kernel size to use is one of the more interesting and pressing issues that must be clarified before our method can be widely applied. These networks have intentionally been kept as simple as possible in order to facilitate comparison, to enhance interpretability of results, and to keep the focus upon the impact of utilizing our mapping algorithm in the deep learning process and not upon exotic CNN architectures. (For a brief discussion of the impact that more sophisticated architectures have upon classification performance when training on data encoded via our method, see the K-Ras / H-Ras results in section 4.) The final configurations for the three CNN variants are summarized in table 1.

\begin{table}[ht]
\centering
\begin{tabular}[t]{ccc}
\toprule
3D CNN & 2D CNN & 1D CNN \\
\midrule
Input (64x64x64) & Input (512x512) & Input (1x262144) \\

Conv3D - stride (2x2x2) & Conv2D - stride (3x3) & Conv1D - stride (9) \\

MaxPool3D - stride (2x2x2) & MaxPool2D - stride (3x3) & MaxPool1D - stride (9) \\

Conv3D - stride (2x2x2) & Conv2D - stride (3x3) & Conv1D - stride (9) \\

MaxPool3D - stride (2x2x2) & MaxPool2D - stride (3x3) & MaxPool1D - stride (9) \\

Dense (128) & Dense (128) & Dense (128) \\

Dropout (0.5) & Dropout (0.5) & Dropout (0.5) \\

Output & Output & Output \\
\bottomrule \\
\end{tabular}
\caption{Base network definitions for 3D, 2D, and 1D networks. Different kernel sizes were used for convolutional layers and the results of those variations are detailed in the results sections below.}
\label{table:1}
\end{table} 

 Throughout our experiments, kernel sizes for 2D and 1D networks were chosen to approximate as closely as possible the number of voxels contained within the kernels selected in their 3D counterparts. The selection of 3D CNN kernel sizes was informed by a general understanding of the likely size that important features might be rendered at given the $64^3$ window size for both the ModelNet and K-Ras / H-Ras data cases. For insight into our team's ongoing research into determining optimal CNN hyperparameter selection for training with our data encodings, see the discussion and future work sections.  

\section{Benchmarking}
\label{case-modelnet}
\subsection{The ModelNet10 Dataset}

\begin{figure}[h]
    \centering
    \includegraphics[width=5cm]{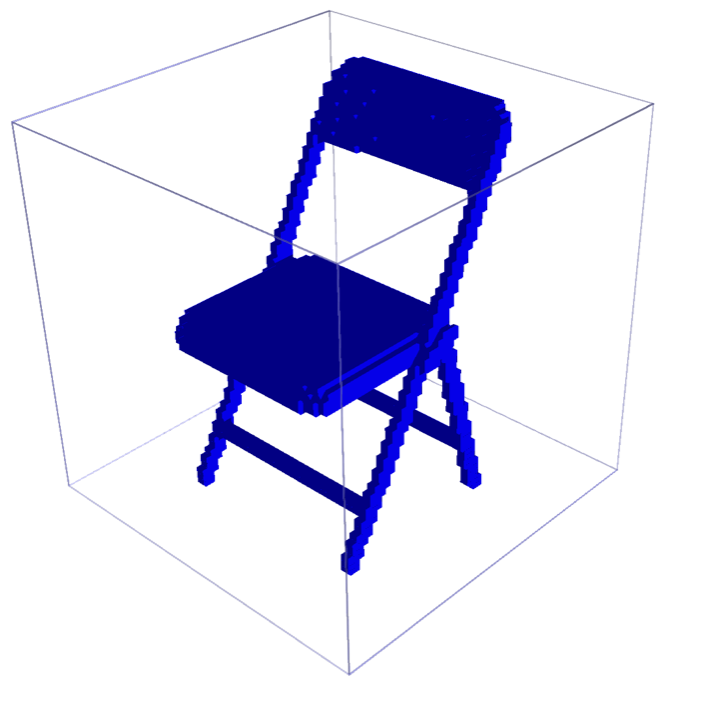}
    \includegraphics[width=7.5cm]{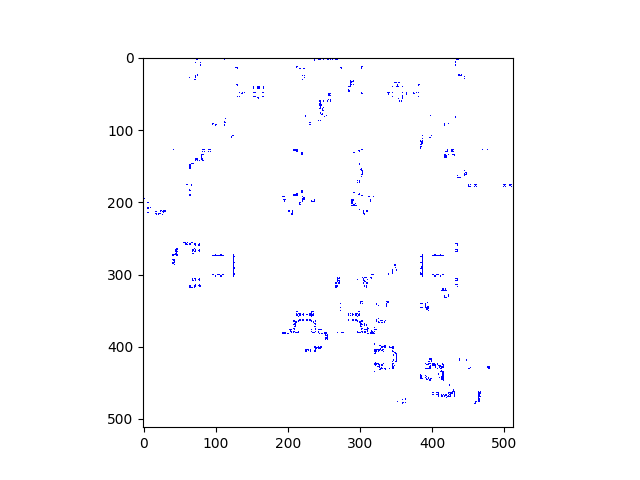}
    \caption{A 3D rendering of a folding chair model discretized in $64^3$ window space on the left and the corresponding 2D encoding of the same model manifesting as a 512x512 pixel image on the right.}
\end{figure}

In order to test and refine our method, we made use of the ModelNet 3D object model benchmarking dataset, which is maintained and sponsored by Princeton \cite{wu20153d}. ModelNet was intended to provide researchers with a clean dataset composed of multiple richly-populated categories of high-quality 3D CAD models of simple objects for use in machine vision and deep learning experiments. The models in this set are purely geometric (meaning no information is encoded in the models other than their structure), and all models have been manually aligned by the maintainers of the dataset so that they have a common orientation. In total, the ModelNet10 set contains approximately 5,000 examples split evenly among ten everyday object classes, including models such as tables and chairs. Figure 3 provides a 3D rendering of an archetypal ModelNet10 chair model discretized in a $64^3$ window space on the left along with its corresponding 2D encoding on the right.The dataset was randomly shuffled into training, validation and test sets using a 70 / 10 / 20 split.   

\subsection{ModelNet10 Classification Results}

The table below details the final test loss and accuracy of networks trained across all versions of the ModelNet10 dataset.

\begin{table}[h]
\centering
\begin{tabular}[t]{cccc}
\toprule
CNN Dimension & Kernel & Test Loss & Test Accuracy\\
\midrule
 & 4x4x4 & 0.383 & 92.7\% \\

3D & 5x5x5 & 0.343 & 92.8\% \\

 & 6x6x6 & 0.414 & 90.9\% \\
\midrule
 & 8x8 & 0.381 & 93.2\% \\

2D & 11x11 & 0.402 & 92.4\% \\

 & 15x15 & 0.439 & 91.5\% \\
\midrule
 & 1x64 & 0.449 & 90.9\% \\

1D & 1x121 & 0.478 & 92.6\% \\

 & 1x225 & 0.421 & 91.8\% \\
\bottomrule \\
\end{tabular}
\caption{ModelNet10 classification accuracy across CNN architectures with varying kernel sizes and data representations. All networks were trained for 20 epochs.}
\label{table:2}
\end{table}

The ModelNet maintainers track the current reported performance of the various approaches to model classification submitted by the community and, as of late January 2018, the top-ranking purely volumetric CNN-based approach is LightNet \cite{LightNet}, which achieves a classification accuracy of 93.94\% on the ModelNet10 dataset. LightNet is, as its name implies, a lightweight volumetric CNN architecture that makes use of approximately 300,000 trainable parameters and a relatively unique configuration incorporating up to 9 fully-connected "dense" layers of 128 neurons each. Although the networks that we used in our experiments express architectures more similar to that of a popular early volumetric CNN called VoxNet \cite{maturana2015voxnet} than to LightNet, they are similar to LightNet in the sense that they improve upon the performance of VoxNet while containing at most a third of that network's ~900,000 trainable parameters. Some of the variants of our networks are actually nearly 50\% smaller even than LightNet (in terms of total trainable parameter counts) while still besting the performance of VoxNet on the ModelNet10 dataset. 

For comparison purposes, it is worth noting that the highest accuracy reported thus far on the ModelNet10 leaderboard comes from an ensemble learning system, Brock et al.'s VRN Ensemble (97.14\%) \cite{VRN_ENSEMBLE}. Among the other successful approaches published to date are Klokov and Lempitsky's point-cloud-based Deep Kd-Networks (94\%) \cite{DBLP:journals/corr/KlokovL17}, Hegde and Zadeh's multi-representation FusionNet (93.11\%) \cite{2016arXiv160705695H}, Sedaghat et al.'s orientation-boosted ORION (93.8\%) \cite{DBLP:journals/corr/AlvarZB16}, and Wu et al.'s probabilistic 3D-GAN (91\%) \cite{2016arXiv161007584W}, none of which rely exclusively upon non-augmented volumetric datasets, and some of which make no use of volumetric data whatsoever. 

Interestingly, the classification accuracy levels attained when training our 3D, 2D, and 1D networks on the various representations of our data were comparable to one another, with the 8x8-kernel 2D CNN variant demonstrating a slight advantage over the others. These results show that a variety of CNN architectures are capable of learning from data encoded via our mapping algorithm.

\subsection{ModelNet10 Computational Performance}

The table below describes the variation in performance across the various CNN architectures trained on the ModelNet10 dataset.

\begin{table}[h]
\centering
\begin{tabular}[t]{cccc}
\toprule
CNN Dimension & Kernel & Parameters & Updates Per Second \\
\midrule
 & 4x4x4 & 179,658 & 39.5 \\

3D & 5x5x5 & 244,074 & 40.0 \\

 & 6x6x6 & 262,346 & 39.3 \\
\midrule
 & 8x8 & 171,466 & 100.8 \\

2D & 11x11 & 231,658 & 87.9 \\

 & 15x15 & 304,618 & 77.9 \\
\midrule
 & 1x64 & 228,810 & 92.6 \\

1D & 1x121 & 284,906 & 85.7 \\

 & 1x225 & 390,634 & 76.2 \\
\bottomrule \\
\end{tabular}
\caption{Performance variation among CNN architectures for the ModelNet10 classification task on one Nvidia K20. Total trainable parameter counts provided per architecture and kernel size. Average training updates per second with batch size of 1.}
\label{table:3}
\end{table}

To measure the computational efficiency of our approach we measured the number of network updates taken per second during training for each of the three versions of the ModelNet10 dataset that we chose to explore, namely, the native 3D data and encoded 1D and 2D versions of the same. The performance results associated with each network configuration, along with the number of trainable parameters for each network, are detailed in table 3.

Even given the mono-channeled nature of the ModelNet data, which is inherently sub-optimal for demonstrating our system's potential channel-carrying and processing advantages, the speedup associated with leveraging a 2D convolution operation versus a 3D one is readily apparent. Our results demonstrate a speedup in per-second network training updates of 2.5 times when using a 2D network with kernel size of 8x8 versus all of the 3D CNN architectures tested. This 2D network configuration also had the highest test accuracy of all the network configurations explored, indicating that data encoded via our mapping algorithm support both high classification performance and high computational efficiency. This trend becomes more visible when data consisting of multiple channels of information are considered, as in K-Ras / H-Ras classification task explored in section 4.

One curious observation our team encountered was the disparity between the performance of 2D and 1D CNNs. Although the 1D convolution procedure might theoretically be expected to operate at a time complexity one power lower than that of its 2D counterpart our experiments show that 2D CNN performance is in fact better than in the 1D case. At this time, we do not have a satisfactory explanation for this phenomenon, but we hypothesize that it may be attributable to implementation details within the TensorFlow library that we have not reviewed. 

\section{Protein Branch Classification via Ras Family Homologes}
\label{case-kras_hras}

\subsection{Ras Structure and Machine Learning}

Here we use our method to classify between sequentially-similar members of the Ras protein family based upon very limited training data. Our primary goal in approaching this classification task is to demonstrate the viability of our data encoding methodology as a means to enable a deep learning-powered exploration of the structural characteristics of complex biomolecules by making efficient use of available data and potentially enabling the high-resolution visualization of learned features.

The Ras subfamily of proteins, with H-Ras, K-Ras, and N-Ras as its canonical members, are of great interest in cancer research \cite{colicelli2004}. There is more than 85 percent homology between the amino acid sequences of these proteins, with their main differences concentrated in the C-terminal hyper-variable region. We do not make use of this highly-variable region for our work, meaning that the data classes used to train our CNNs are highly similar to one another and are differentiated only by extremely minute features. K-Ras is of particular interest because its mutation is frequently coincident with the development of human cancers \cite{McCormick1797}. K-Ras is known to be the most frequently mutated of the three Ras family branches and the most common in human cancer \cite{Prior2457}, and for these reasons the understanding of this protein has been highlighted as particularly important to the efforts of the National Cancer Institute's (NCI) Ras Initiative. Ras proteins have posed numerous challenges to researchers in a variety of directions for decades. For instance, they are widely considered to be undruggable due to their lack of obvious cavities or pockets on their small globular surfaces \cite{shokat2016}. 

There are significant functional differences between the various Ras branches despite their highly homologous primary protein sequences, and although there has been extensive research conducted to investigate the cause and character of this phenomenon, the simple fact is that, generally speaking, neither the mechanism by which a protein conforms into its unique 3D structure nor that structure's impact on the protein's biological functionality are fully understood. Thus, a detailed 3D structural characterization of Ras family members, and the resulting ability to classify K-Ras and its related mutants based upon that structural knowledge, could provide pivotal insights to the functional differences between the members of the Ras family, and even between the various members of the K-Ras subfamily itself. It is our hope that successful processing of 3D Ras structures using CNNs may help to illuminate methods of identifying potential binding sites, as well as inhibitor candidates. 

\subsection{K-Ras and H-Ras Dataset}

\begin{figure}[ht]
    \centering
    \includegraphics[width=6.5cm]{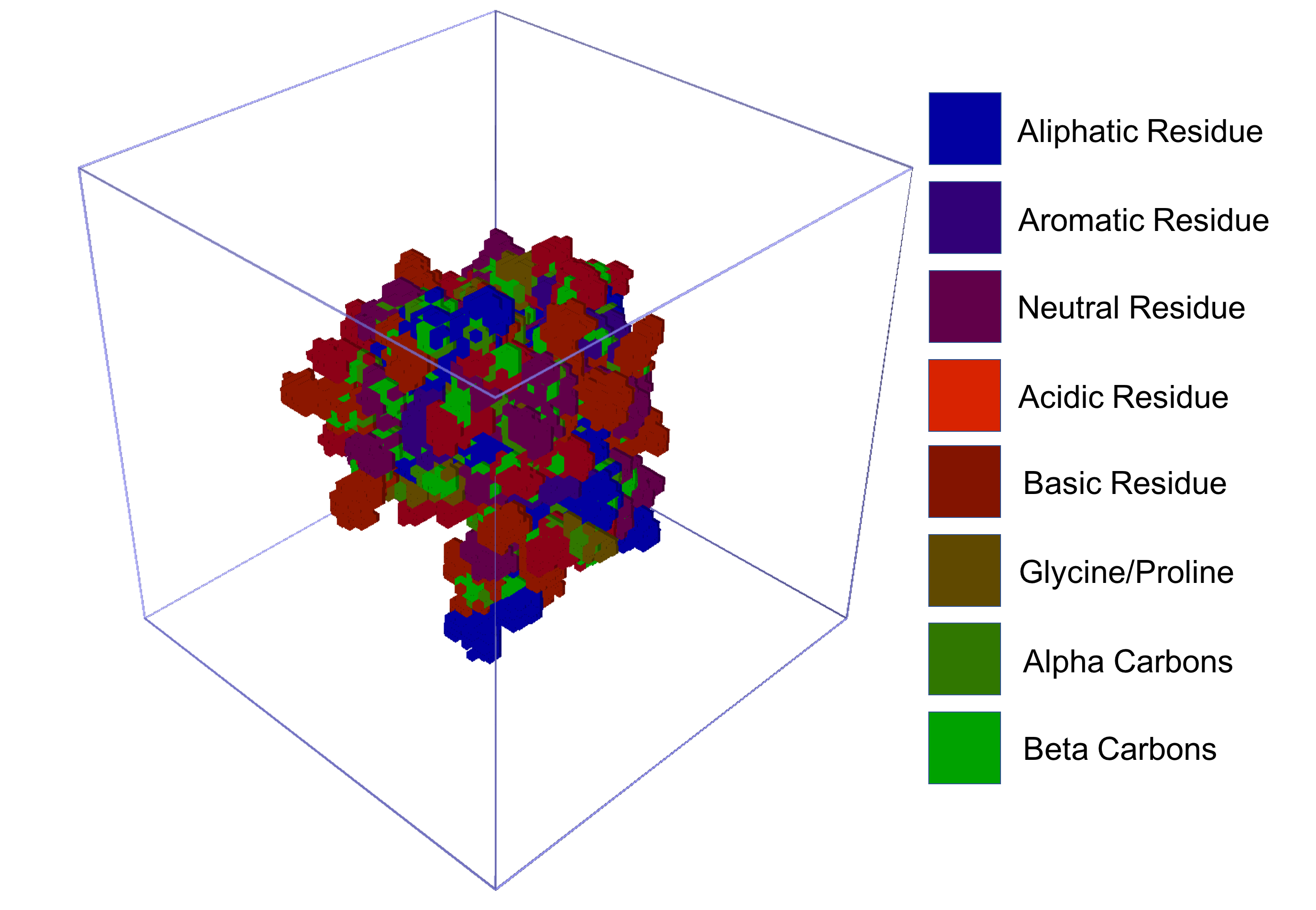}
    \includegraphics[width=7cm]{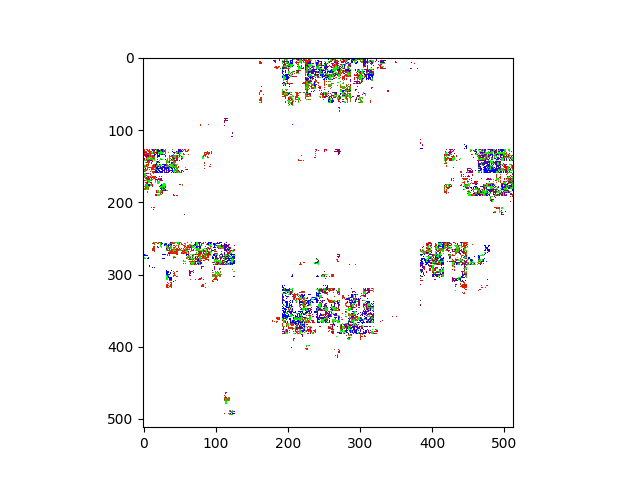}
    \caption{A 3D rendering of an 8-channel Ras model (ID 1aa9) discretized in $64^3$ window space on the left and the corresponding 2D encoding of the same model on the right. The colors blue, purple, magenta, light red, dark red, brown, dark-green, and light-green correspond to aliphatic residues,aromatic residues, neutral residues, acidic residues, basic residues, glycine/proline, alpha carbons, and  beta carbons, respectively.}
\end{figure}

We designed a binary classification task between the K-Ras and H-Ras subfamilies. Structural data of 239 unique protein chains was sourced from the RCSB Protein Data Bank (PDB) \cite{doi:10.1093/nar/28.1.235}, with 161 chains belonging to the H-Ras subfamily, and 78 chains belonging to the K-Ras subfamily. To begin, we created binary volumetric representations of each protein chain using atomic coordinate data. Prior to voxelization, the principle directions of each protein chain were determined using principle component analysis (PCA) on the chain's atomic coordinates similar to the approach taken by the EnzyNet team \cite{2017arXiv170706017A}. Each protein chain was then aligned according to these axes in order to provide a comparison frame among all chains. The origin of each model was taken to be its center of mass. A space-filling representation of each atom was generated based upon its van der Waal radius, yielding a rich overall model for each chain in the dataset. Finally, each chain was discretized within a $64^3$ window, resulting in a per-voxel resolution of 1 cubic angstrom. This resolution was selected to allow for all K-Ras and H-Ras chains to fit within the defined window space when rendered, as well as to allow for a CNN trained upon such renderings to learn features at a biologically-meaningful scale.

The final volumetric representations of protein structures comprising our dataset also include 8 channels of information beyond the purely geometric voxel occupancy data described above. These channels were selected based upon their likelihood of describing important differences between K-Ras and H-Ras at a level of detail adequate to overcome the two proteins' extreme structural homology. Specifically, the following channels of information were incorporated into our representations on a per-atom basis: alpha and beta carbons (useful for describing a chain's backbone composition, as well as the orientation of side chains), atoms belonging to hypdrophobic amino acids as either aliphatic or aromatic residue members, atoms belonging to charged amino acids as either neutral, acidic, or basic residue members, glycine, and proline. This 3D dataset was then converted into both 2D and 1D formats via the methodology described in section 2, providing us with three different representations of the same structural protein data to train with. Figure 3 shows the 3D rendering of an 8-channel Ras protein discretized in a $64^3$ window space on the left along with its corresponding 2D encoding on the right. Each dataset generated was randomly shuffled and separated into training, validation, and test sets using a 70/10/20 split. 

Although it is quite possible to make use of a relatively powerful rotation-based dataset augmentation strategy when dealing with structural data for which no "native" orientation exists, such as with proteins, the value of a machine learning system is closely related to its ability to train on limited data. Since the nature of 3D protein datasets - and datasets within the domain of structural biology in general - is to have relatively few examples for each given class, it is important that a system intended for widespread use be able to operate effectively on extremely limited data. For this reason, we opted to gauge the performance of our data encoding methodology and related neural networks without augmenting the number of class elements present in our dataset through random rotations. We will note, however, that is is possible that the use of such a strategy could significantly improve the generalization capability of trained networks if the correct augmentation factor can be found (i.e., not so many that over-fitting occurs).

\subsection{K-Ras / H-Ras Classification Results}

The table below details the final test loss and accuracy of networks trained across all versions of the K-Ras / H-Ras dataset.

\begin{table}[H]
\centering
\begin{tabular}[t]{cccc}
\toprule
CNN Dimension & Kernel & Test Loss & Test Accuracy\\
\midrule
 & 4x4x4 & 0.304 & 87.5\% \\

3D & 5x5x5 & 0.167 & 91.67\% \\

 & 6x6x6 & 0.307 & 83.3\% \\
\midrule
 & 8x8 & 0.188 & 91.67\% \\

2D & 11x11 & 0.201 & 91.67\% \\

 & 15x15 & 0.187 & 91.67\% \\
\midrule
 & 1x64 & 0.275 & 87.5\% \\

1D & 1x121 & 0.232 & 87.5\% \\

 & 1x225 & 0.251 & 87.5\% \\
\bottomrule \\
\end{tabular}
\caption{K-Ras / H-Ras classification accuracy across CNN architectures with varying kernel sizes and data representations. All networks were trained for 20 epochs.}
\label{table:4}
\end{table}

It is worth noting that the task of learning from and classifying between members of the K-Ras and H-Ras branches benefits dramatically from the inclusion of additional information channels in the various data representations used. The generalization abilities of networks of any dimensionality trained on the K-Ras / H-Ras dataset improve from approximately 66\% when mono-channel representations are used to our reported peak of 91.67\% in the 8-channel case. Network loss measurements were reduced from an average of .61 across all CNNs trained on the mono-channeled data to less than half that amount (and in some cases less than a third) for all networks trained on the 8-channeled dataset. Thus the inclusion of these additional channels of information played a key role in successfully training a CNN to discriminate between the two protein classes. Similar cases where additional data richness would critically impact system performance can be imagined elsewhere in structural biology, computational chemistry, or even cosmology.

Furthermore, the results reported in this and previous sections do not represent the maximum attainable classification performance of CNNs trained on data encoded via our mapping algorithm. For instance, an uncomplicated 13-layer 2D CNN architecture based upon the design principles laid out in the SimpleNet publication \cite{DBLP:journals/corr/HasanPourRVS16} is capable of classifying between 2D Ras encodings with 99\% accuracy after only a handful of training epochs, albeit at the cost of increasing the total parameter count over our simpler architectures by an order of magnitude to approximately 5 million. Similarly, performance on the raw 3D data can be improved by a large margin by using a slightly deeper 3D CNN architecture with greater kernel sizes and additional, larger densely-connected layers. One such 3D network trained by our team was able to distinguish between K-Ras and H-Ras examples with 99\% accuracy after only two training epochs, although this network was significantly heavier even than the SimpleNet-based 2D one, with nearly 11 million trainable parameters in total.

The key takeaway of the above results, then, should be that there exists a wide variety of different CNN architectures that are capable of effectively learning from complex structural data encoded into $N<3$ dimensions via our mapping algorithm, and that there are legitimate cases where the successful training of a CNN can depend upon the channel carrying capacity of a data representation scheme. The fact that simple CNN architectures such as those used in our experiments are capable of learning to classify between extremely similar, complex data classes based upon very limited training data encoded via our mapping algorithm implies that the encoded representation of said data reliably expresses essential class features in a learnable form. It seems highly likely that the development of 1D and 2D CNN hyperparameter-selection heuristics for training with our data encodings, as well as the further improvement of our encoding algorithm itself, will yield strong results on problems in domains characterized by complex structural data (see the section titled "Future Work" for more details). 

\subsection{K-Ras / H-Ras Computational Performance}

The performance results associated with each network configuration, along with the number of trainable parameters for each network, are detailed in table 5. This case differs significantly from the ModelNet10 case detailed in section 3 due to the multi-channel nature of the K-Ras / H-Ras dataset. 

There are several important components of our results in this case that deserve to be discussed in some detail, especially in contrast to those attained when training with the ModelNet10 dataset. Of primary importance when evaluating these performance results is the comparison between the updates per second measurements for 3D, 2D, and 1D networks trained on the mono-channeled ModelNet10 data and the 8-channeled K-Ras / H-Ras data.

When trained on this 8-channel data, our 3D CNNs were able to process only 78\% as many updates per second on average as the same networks trained on the mono-channel ModelNet data. However, when 2D CNNs were trained on 2D encodings of this same structural data, they were able to process an average of 88\% as many per-second updates as 2D encodings of the mono-channel ModelNet data. Furthermore, in both the ModelNet and K-Ras / H-Ras cases, networks operating in 2D (i.e., trained upon 2D representations of 3D structural data encoded via our method) were able to process over twice as many per-second updates as their 3D counterparts.

This experiment provides a good demonstration of our encoding algorithm's potential channel-carrying and processing advantages, since it highlights the speedup associated with leveraging a 2D convolution operation versus a 3D one in the context of complex structural data. It clearly demonstrates that the slowdown incurred by increasing the number of channels contained in training data is greater in the 3D case than in 2D. Importantly, this 2D CNN speed advantage is accomplished while still retaining classification accuracy levels on-par with those obtained from slower 3D CNNs. 

The table below describes the variation in performance across the various CNN architectures trained on the K-Ras / H-Ras dataset.

\begin{table}[H]
\centering
\begin{tabular}[t]{cccc}
\toprule
CNN Dimension & Kernel & Parameters & Updates Per Second \\
\midrule
 & 4x4x4 & 192,962 & 32.6 \\

3D & 5x5x5 & 271,042 & 32.6 \\

 & 6x6x6 & 309,698  & 27.5 \\
\midrule
 & 8x8 & 184,770 & 81.5 \\

2D & 11x11 & 257,730 & 75.8 \\

 & 15x15 & 353,986 & 75.8 \\
\midrule
 & 1x64 & 242,114 & 48.2 \\

1D & 1x121 & 310,978 & 45.8 \\

 & 1x225 & 440,002 & 45.0 \\
\bottomrule \\
\end{tabular}
\caption{Performance variation among CNN architectures for the K-Ras / H-Ras classification task on one Nvidia K20. Total trainable parameter counts provided per architecture and kernel size. Average training updates per second with batch size of 1.}
\label{table:5}
\end{table}

Finally, just as the case with ModelNet10, computational training performance in the 1D case was found to be worse than expected, possibly due to underlying implementation details in TensorFlow that our team have not reviewed. In particular, our 1D CNNs training on linear representations of 3D K-Ras / H-Ras data were only able to process 55\% of the updates per second recorded for the mono-channel ModelNet case, on average.

\section{Discussion}

We have proposed an algorithm that maps 3D structural data down to 2D and 1D, allowing it to be efficiently processed by traditional (i.e., non-volumetric) CNNs. The mapping of the structural data clearly preserves a proportion of the structure's clustered semantic regions large enough to enable the training of lower-dimensioned CNNs. We evaluated the classification performance of 3D, 2D, and 1D CNNs on the ModelNet10 benchmark and found there to be comparable performance between the three network types for that task. Our results indicate the strong potential for our data encoding algorithm to play a critical role in improving the performance of pre-existing systems. It is possible, for example, that data encoded via our algorithm could be used in place of the relatively unsophisticated 2D projections currently being used by the multi-representation FusionNet, potentially improving that system's already strong performance by enhancing the richness of its available data representations.

It is clear based upon the experiments we have conducted that the process of encoding high-dimension datasets into lower-dimension representations via a space-filling curve-based mapping algorithm is an effective way to reduce the time cost associated with training CNNs on such data. Our results indicate that CNNs trained on encoded 2D representations of 3D structural data are able to classify between the 3D structures to a degree of accuracy that matches, and in some cases even surpasses, that of a 3D CNN. Furthermore, it has been shown that by mapping 3D structural data into 2D far more data channels can be utilized. This is a very important feature that points toward our data encoding method being potentially applicable to domains such as computational chemistry and structural biology, where high-resolution, multi-channel representations of various data are very much in demand. The fact that our system supports the incorporation of large numbers of information channels into data representations means that it can scale beyond basic applications, such as ModelNet10's everyday object recognition task, and be leveraged against classification problems for which the only way to distinguish between class members is through the processing of large numbers of additional descriptors, as was the case with the homologous K-Ras and H-Ras protein subfamilies. 

Furthermore, although there remains a significant amount of work to be done in order to understand the exact characteristics of our mapping algorithm, it is clear that the process supports encoding data between arbitrary native and target dimensions. This raises interesting possibilities for the construction of applications that process data of more than three dimensions as 3D, 2D, or 1D data grids. Another intriguing possibility lies with the compression of data, where, for instance, 2D data encodings might be generated from 3D structural data before being compressed using standard (or customized) image compression algorithms to further increase training performance. Compression might also be possible in the case of 1D representations of our 3D data, where the large number of unoccupied elements in the vector (which would almost always be a huge percentage of the total vector size due to the sparse nature of most structural data) could potentially be compressed by a factor of 10-to-one or more, yielding relatively small representations that would theoretically be very efficient to work with while still preserving mapping reversibility for visualization purposes.

An interesting question that we have not yet addressed is what effect the size of a 3D model rendered within a given window space has upon a trained network's ability to recognize the model and if there is a benefit associated with model size normalization. Specifically, we are unsure of the minimum size that a model may be rendered at within a given window while still being reliably recognizable by trained 3D networks such as those that we made use of. This question takes on additional importance when the aforementioned possibility of image and vector compression is considered. Preliminary results obtained by our team indicate that 2D CNNs trained on 512x512 pixel encoded versions of K-Ras and H-Ras protein models rendered in a $64^3$ window space are able to distinguish between the two classes with high degrees of accuracy and generalization capability even when the 2D representations are downsampled to a size of 64x64 pixels. Detrimental effects caused by compression appeared to be minimal, and the improvement in computational performance in both training and inference phases was substantial. But while such an approach may provide a boost in performance, the negative implications it carries for the interpretability of results and feature identification are significant. It is our intention to explore the issues of model size normalization and image and vector downsampling in a more thorough manner in the future.  

On a final, technical note, much of our research was carried out on an Nvidia DGX-1 at Oak Ridge National Lab's Leadership Computing Facility. The DGX-1 is a deep learning-optimized supercomputer equipped with eight Nvidia Tesla P-100 GPU cards. Along with the DGX-1, we also used XSEDE's Comet located at SDSC, which boasts four P-100 GPU per node. NERSC, located at Lawrence Berkeley National Lab, provided access to the Cori supercomputer, which was used primarily for data generation and preprocessing due to its CPU-based architecture. Although Cori has recently been reported to perform well in a full-system deep learning task using the Caffe framework \cite{2017arXiv170805256K}, we were not able to make large-scale use of the system in our own TensorFlow-based CNN training, since at this time TensorFlow does not reliably scale beyond a single Haswell / KNL node on that system. Berkeley Lab's Nvidia K20-powered CAMERA GPU cluster was utilized by our team during the final testing and tuning phases of our network design process and was used to generate the results that appear in this paper. All of these results were attained using the Keras 2.0 API on top of the TensorFlow 1.2 computation backend. 

\section{Future Work}

Our team is currently engaged in several different efforts designed to expand the interpretability and applicability of the various components of our mapping algorithm and concomitant CNN training processes. Among the more interesting and potentially-fruitful avenues we are exploring are applications in enzyme classification, protein-ligand binding prediction, and computationally-generated protein model scoring.

One of the more important questions that must be answered before our methodology can be widely applied by other researchers is how to determine the appropriate set of hyperparameters when designing CNNs to train upon data encoded via our mapping algorithm. The 2D deep learning vision community benefits from a well-established set of hyperparameter-selection heuristics that have been built up over time based upon the accumulated experiences of many thousands of researchers, as well as from rigorous mathematical analyses. In order to allow for the straightforward application of our technology to complex structural domains we will need to provide a set of actionable CNN hyperparameter-selection heuristics tailored specifically for maximizing the benefit of data encoded by our method. Our team is currently exploring the impact of various hyperparameters upon CNN performance in the context of our data encoding algorithm, and we intend to make the results of this exploration public in future versions of this paper to support the adoption of our technology among the community.  

An area of particular excitement is related to the interpretation of CNN operations and results, and the extraction of class features. Since our mapping algorithm is bijective, visualization of learned class features via the reversal of saliency maps into 3D from 1D and 2D is possible. For example, this means that networks trained on 2D encodings of 3D protein structures - i.e., networks learning dispersed, 2D versions of the protein's critical 3D features - can be used to generate 2D attention maps that, when run through our mapping algorithm "backward", will generate 3D attention maps highlighting the 3D features seized-upon by the 2D network. This capability, along with the others put forth in this paper, potentially offers a strong argument for the use of our technology as one component of a cost-effective methodology for the rapid exploration of complex structural domains using pre-existing deep learning tools. 

\section{Acknowledgements}

This work was supported in part by the U.S. Department of Energy, Office of Science, Office of Workforce Development for Teachers and Scientists (WDTS) under the Visiting Faculty Program (VFP) and Berkeley Lab Undergraduate Faculty Fellowship (BLUFF) Program. It was also supported by the Director, Office of Science, Office of Basic Energy Sciences, of the U.S. Department of Energy under Contract No. DE-AC02-05CH11231. It was in part supported by the NSF Blue Waters Project under the Student Internship Program.

The authors would like to acknowledge Prof. Nelson Max of University of California, Davis and Dr. Peter Zwart of Lawrence Berkeley National Laboratory for their constructive feedback.

Computing allocations were provided through:
  \begin{itemize}
    \item LBNL/NERSC Repo M1532
    \item ORNL/OLCF CSC 260 Protein Feature Extraction from 3D Structure using CNN
    \item XSEDE CIE-140013 Advancing Computational Education in Small Colleges
  \end{itemize}

\bibliographystyle{unsrt}

\small

\bibliography{main}

\end{document}